\documentclass[preprint,12pt]{elsarticle}

\usepackage{amssymb}
\usepackage{multirow}
\usepackage{makecell}
\usepackage{booktabs}
\usepackage{amsmath}
\usepackage{float}
\usepackage[hyphens]{url}
\usepackage{hyperref}
\usepackage[numbers]{natbib}
\hypersetup{colorlinks=true,breaklinks=true}
\journal{XXX}

\begin{document}

\begin{frontmatter}

%% Title, authors and addresses

%% use the tnoteref command within \title for footnotes;
%% use the tnotetext command for theassociated footnote;
%% use the fnref command within \author or \address for footnotes;
%% use the fntext command for theassociated footnote;
%% use the corref command within \author for corresponding author footnotes;
%% use the cortext command for theassociated footnote;
%% use the ead command for the email address,
%% and the form \ead[url] for the home page:
%% \title{Title\tnoteref{label1}}
%% \tnotetext[label1]{}
%% \author{Yuqing Wang\corref{cor1}\fnref{label1}}
%% \ead{wang603@ucsb.edu}
%% \ead[url]{home page}
%% \fntext[label2]{}
%% \cortext[cor1]{}
%% \affiliation{organization={},
%%             addressline={},
%%             city={},
%%             postcode={},
%%             state={},
%%             country={}}
%% \fntext[label3]{}

\title{An Empirical Study on the Robustness of the Segment Anything Model (SAM)}

%% use optional labels to link authors explicitly to addresses:
%% \author[label1,label2]{}
%% \affiliation[label1]{organization={},
%%             addressline={},
%%             city={},
%%             postcode={},
%%             state={},
%%             country={}}
%%
%% \affiliation[label2]{organization={},
%%             addressline={},
%%             city={},
%%             postcode={},
%%             state={},
%%             country={}}

\author[label1]{Yuqing Wang}
\author[label2]{Yun Zhao}
\author[label1]{Linda Petzold}

\affiliation[label1]{organization={Computer Science Department, University of California, Santa Barbara},%Department and Organization 
            country={United States}}
\affiliation[label2]{organization={Meta Platforms, Inc.},%Department and Organization 
            country={United States}}

\begin{abstract}
The Segment Anything Model (SAM) is a foundation model for general image segmentation. Although it exhibits impressive performance predominantly on natural images, understanding its robustness against various image perturbations and domains is critical for real-world applications where such challenges frequently arise. In this study we conduct a comprehensive robustness investigation of SAM under diverse real-world conditions. Our experiments encompass a wide range of image perturbations. 
Our experimental results demonstrate that SAM's performance generally declines under perturbed images, with varying degrees of vulnerability across different perturbations. By customizing prompting techniques and leveraging domain knowledge based on the unique characteristics of each dataset, the model's resilience to these perturbations can be enhanced, addressing dataset-specific challenges. This work sheds light on the limitations and strengths of SAM in real-world applications, promoting the development of more robust and versatile image segmentation solutions.
\end{abstract}

\begin{keyword}
%% keywords here, in the form: keyword \sep keyword

%% PACS codes here, in the form: \PACS code \sep code

%% MSC codes here, in the form: \MSC code \sep code
%% or \MSC[2008] code \sep code (2000 is the default)
Image segmentation \sep Segment Anything Model \sep Model robustness evaluation \sep Prompting methods \sep Domain-specific analysis

\end{keyword}

\end{frontmatter}

%% \linenumbers

%% main text
\section{Introduction}
\label{intro}
The emergence of foundation models~\cite{bommasani2021opportunities} has brought about a significant paradigm shift in various domains, as these models achieve impressive results due to their extensive pre-training on massive datasets and their remarkable generalization abilities across a wide array of downstream tasks. Foundation models are designed to serve as a common backbone that can be fine-tuned or adapted to numerous tasks, making them highly versatile and efficient. In the natural language processing (NLP) domain, the Generative Pre-trained Transformer (GPT)~\cite{brown2020language} by OpenAI has set new benchmarks in various language tasks and paved the way for successful commercial applications like ChatGPT~\cite{openai2023gpt4}, known for its real-time and coherent language generation and user interaction. However, in the realm of computer vision, the quest for equally powerful and adaptable foundation models continues, as researchers strive to overcome the unique challenges and complexities associated with the development of these models in the visual domain.

The development of CLIP~\cite{radford2021learning}, a model that effectively combines image-text modalities, demonstrated the potential for zero-shot generalization to previously unseen visual concepts. However, due to the limited availability of comprehensive training data, its generalization capacity for vision tasks remains constrained compared to NLP models. More recently, Meta AI Research introduced the Segment Anything Model (SAM)~\cite{kirillov2023segment}, a versatile and promptable model capable of segmenting any object in images or videos without the need for additional training, a process known as zero-shot transfer in the vision community. Distinct from previous models, SAM serves as the first foundation model in computer vision trained on the extensive SA-1B dataset, which consists of over 11 million images and one billion masks. SAM is engineered to deliver accurate segmentation results using a variety of prompts such as points, boxes, or a combination of both, and has undeniably exhibited strong generalization across diverse images and objects. This success has opened up new possibilities and opportunities for diverse image analysis and understanding~\cite{ma2023segment, ji2023sam, zhou2023can, deng2023segment}.

The impressive generalization capabilities of SAM warrant further investigation of its robustness when confronted with a wide variety of real-world scenarios and imaging modalities. Images captured in diverse environments are subject to factors such as low-light settings, noise, blurring, and compression artifacts. Moreover, various imaging modalities, including satellite, radiology, and ultrasound images, present unique challenges for segmentation models. Satellite images can be low-resolution and noisy, while radiology images can be complex and contain various tissues. Given that it is impossible for a dedicated pre-training dataset to cover the full spectrum of real-world situations, evaluating SAM's adaptability and generalization under these different applications becomes crucial~\cite{kamann2020benchmarking}.

In light of this situation, our study conducts a comprehensive evaluation of SAM's robustness across a wide range of image perturbations and domains, including various noise levels, blurring effects, compression artifacts, and imaging modalities such as satellite, radiology, and ultrasound images. We investigate the impact of different prompting techniques on the model's performance and analyze its vulnerabilities and stability under specific perturbations. Our findings reveal that SAM's performance varies significantly across different perturbations, and the choice of prompting method plays a crucial role in enhancing its robustness. Furthermore, we demonstrate the importance of considering the unique characteristics of each dataset when assessing SAM's performance. Through this thorough assessment, we aim to provide a better understanding of SAM's adaptability and generalization capabilities in real-world scenarios.

In summary, our contributions are threefold:
\begin{enumerate}
\item[(1)] To the best of our knowledge, our study provides the first comprehensive analysis of SAM's performance under various perturbations and domains, revealing its strengths and weaknesses in handling diverse real-world situations. We identify specific perturbations that significantly impact the model's resilience, thus uncovering potential areas for improvement.
\item[(2)] By exploring the effects of different prompting methods, such as point, box, and their combinations, on SAM's performance, we demonstrate the importance of tailored prompting strategies in enhancing model resilience. Our findings emphasize the potential benefits of human-in-the-loop interactions in mitigating the impact of perturbations.
\item[(3)] Our dataset-specific analysis reveals the relationship between unique dataset characteristics, such as color, texture, and patterns, and model robustness. This finding suggests that integrating domain knowledge tailored to individual challenges can ultimately enhance the model's resilience against various perturbations.
\end{enumerate}

\section{Related work}
Image segmentation is a crucial component of computer vision, with numerous applications across various domains such as autonomous driving~\cite{zhang2016instance}, medical imaging~\cite{malhotra2022deep, wang2018deepigeos}, video surveillance~\cite{cao2020ship}. It enables the precise identification and localization of objects within images, playing a vital role in processing and interpreting visual data.

Owing to the growing interest in image segmentation, various models have been developed to cater to specific domains. For example, U-Net~\cite{ronneberger2015u} is specifically designed for biomedical image segmentation, utilizing an encoder-decoder architecture to achieve accurate segmentation results. Mask R-CNN~\cite{he2017mask} builds upon the Faster R-CNN~\cite{ren2015faster} object detection model to incorporate instance segmentation capabilities, making it suitable for tasks requiring fine-grained object delineation. DeepLab~\cite{chen2017deeplab} employs atrous convolutions and fully connected conditional random fields to address the semantic segmentation of natural images. Pyramid Scene Parsing Network~\cite{zhao2017pyramid} has been developed for scene parsing tasks, employing pyramid pooling to capture global context information and improve segmentation accuracy. 

Despite the success of these specialized models, there is a growing demand for general-purpose image segmentation models that can be applied across a wide range of applications~\cite{zhang2021k, cheng2021per, cheng2022masked}. SAM~\cite{kirillov2023segment}, as the first foundation model in computer vision, addresses this need by offering a promptable approach to general image segmentation. However, ensuring SAM's effectiveness in real-world applications presents a challenge~\cite{tang2023can, zhou2023can, ji2023segment}, as it necessitates a comprehensive evaluation of its resilience against various image perturbations and domains. With this in mind, our work focuses on investigating the robustness of SAM, examining its strengths and limitations when confronted with a diverse range of perturbations and challenging conditions.

\section{Methods and Experiments}
In this section we outline the methods and experiments conducted to evaluate SAM's robustness under diverse conditions. We begin with a brief overview of the Segment Anything Model (SAM) and the datasets used in our study. Next, we describe our experimental setup, encompassing three major prompting methods and fifteen image perturbation types designed to mimic real-world challenges. Lastly, we present the evaluation metrics employed to assess SAM's performance. 
\subsection{Model Overview}
The Segment Anything Model (SAM) is a foundation model for image segmentation. Drawing inspiration from the concept of “prompting" in NLP~\cite{wei2022chain, wang2023large}, SAM is a promptable model that supports various types of visual and textual prompts. SAM consists of three main components: an image encoder, a flexible prompt encoder, and a fast mask decoder. The image encoder is built upon a pre-trained Vision Transformer (ViT)~\cite{dosovitskiy2020image} that has been minimally adapted to process high-resolution inputs. In our experiments, we use the default ViT-H image encoder. The prompt encoder accommodates both sparse prompts (points, boxes, text) and dense prompts (masks). The mask decoder efficiently maps the image and prompt embeddings to a mask, utilizing a modified Transformer decoder block with bidirectional cross-attention between prompt and image embeddings. Finally, an upsampling step and an MLP are employed to compute the mask foreground probability at each image location. For a more detailed description of SAM's architecture and training procedure, we refer readers to the original paper~\cite{kirillov2023segment}.

\subsection{Datasets}
\begin{figure*}[htbp]
\centering
\includegraphics[width=\textwidth]{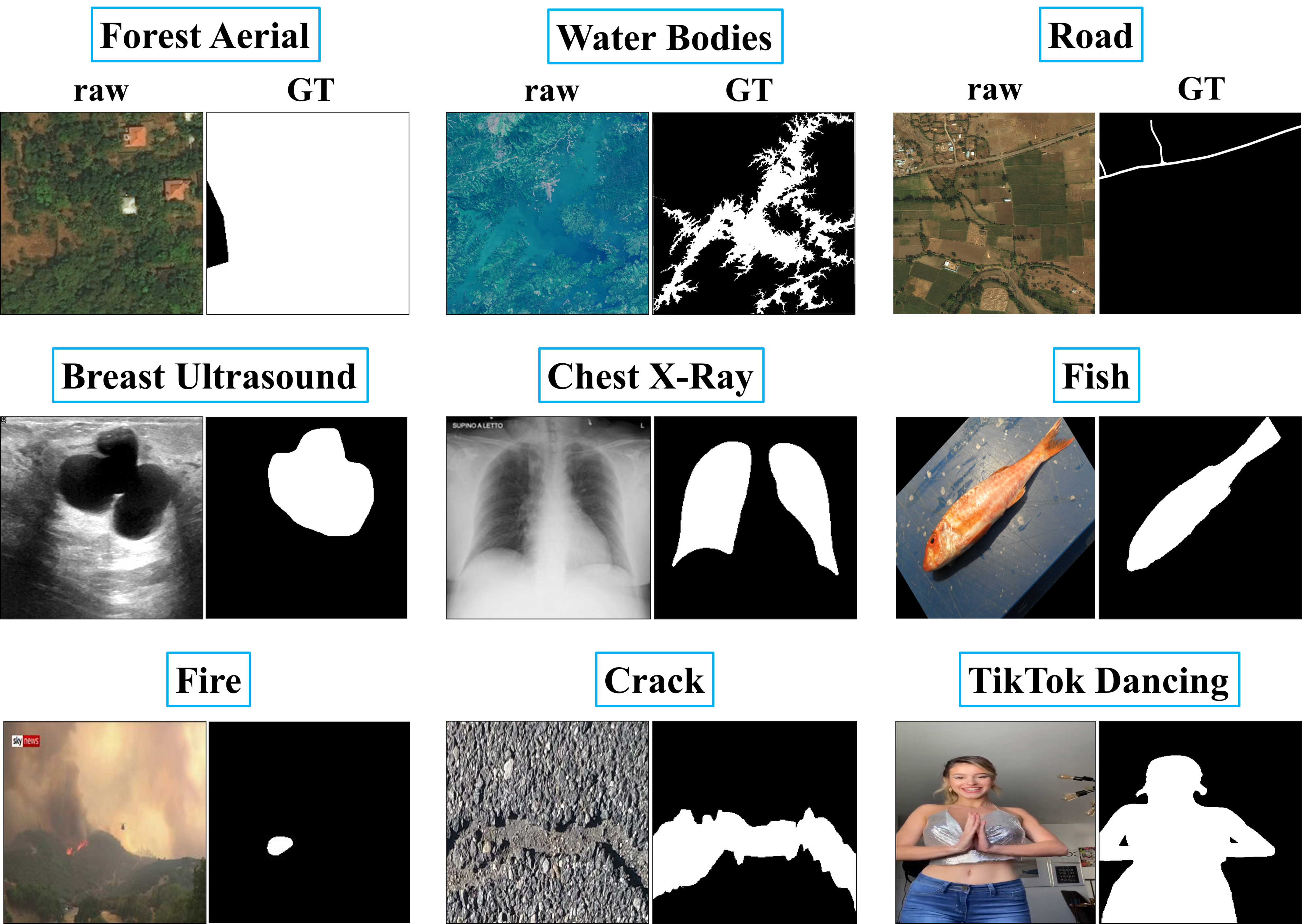}  
\caption{Sample images (raw) and their corresponding ground truth (GT) masks from each dataset.}
\label{fig: sample_images}
\end{figure*}

To thoroughly evaluate the performance and robustness of SAM, we have carefully selected nine datasets that span across distinct imaging conditions and pose various segmentation challenges. The Remote Sensing and Geographical category comprises datasets with aerial and satellite imagery, which present hurdles such as diverse resolutions, intricate patterns, and large-scale structures that need to be processed. Medical Imaging datasets, featuring ultrasound and X-ray modalities, often grapple with issues like noise, artifacts, and fluctuating contrasts. The Environment and Natural Phenomena category encompass datasets with a multitude of dynamic elements, such as fish schools with overlapping shapes and fire spread with irregular boundaries, requiring the model to adapt to varying shapes and textures. Finally, the Structural and Human Motion category incorporates datasets with detailed structures like cracks and the intricate motion patterns of human dance, necessitating high precision and the ability to capture subtle nuances. Table~\ref{tab: datasets} presents an overview of the datasets used in our experiments, all of which have binary mask ground truth annotations. To further illustrate the unique characteristics of each dataset, Figure~\ref{fig: sample_images} shows representative raw images along with their corresponding ground truth masks.

\begin{table*}[htbp]
\centering
\caption{Summary of datasets used in the experimental evaluation.}
  \resizebox{\linewidth}{!}{%
\begin{tabular}{cccc}
\hline
\textbf{Category} & \textbf{Dataset} & \textbf{Modality} & \textbf{Num. Images} \\ \hline
\multirow{3}{*}{\makecell{Remote Sensing \\ and Geographical}} & Forest Aerial~\cite{demir2018deepglobe} & Aerial & 5,108 \\
& Water Bodies~\cite{escobar2021satellite} & Satellite & 2,841 \\
& Road Extraction~\cite{demir2018deepglobe} & Satellite & 8,570 \\ \hline
\multirow{2}{*}{Medical Imaging} & Breast Ultrasound~\cite{al2020dataset} & Ultrasound & 780 \\
& Chest X-Ray~\cite{rahman2021exploring} & X-Ray & 18,479 \\ \hline
\multirow{2}{*}{\makecell{Environment and \\ Natural Phenomena}} & Fish~\cite{ulucan2020large} & RGB & 9,000 \\
& Fire~\cite{diversis2021fire} & RGB & 110 \\ \hline
\multirow{2}{*}{\makecell{Structural and Human \\ Motion Analysis}} & Crack~\cite{middha2021crack} & RGB & 11,200 \\
& TikTok Dancing~\cite{jafarian2021learning} & RGB & 100,000 \\ \hline
\end{tabular}}
\label{tab: datasets}
\end{table*}

\subsection{Experimental Setup}
We evaluate the Segment Anything Model within the context of zero-shot learning, examining its robustness against a variety of perturbations and its performance across nine distinct datasets. Our investigation includes three main prompting methods and incorporates fifteen commonly encountered image perturbations.

\subsubsection{Prompting Methods}
We employ three major types of prompting methods for the SAM: point, box, and a combination of point and box prompts. These methods guide the model in its segmentation task by providing varying levels of information about the target object. For point prompting, we explore two variants: single-point and multiple-point, with the choice between them depending on the dataset and the complexity of the object being segmented. The box prompt involves drawing a bounding box around the object, supplying the model with additional spatial information. In the combination method, we integrate both point and box prompts to further enhance the model's understanding of the object's location and shape.

\begin{figure*}[htbp]
\centering
\includegraphics[width=\textwidth]{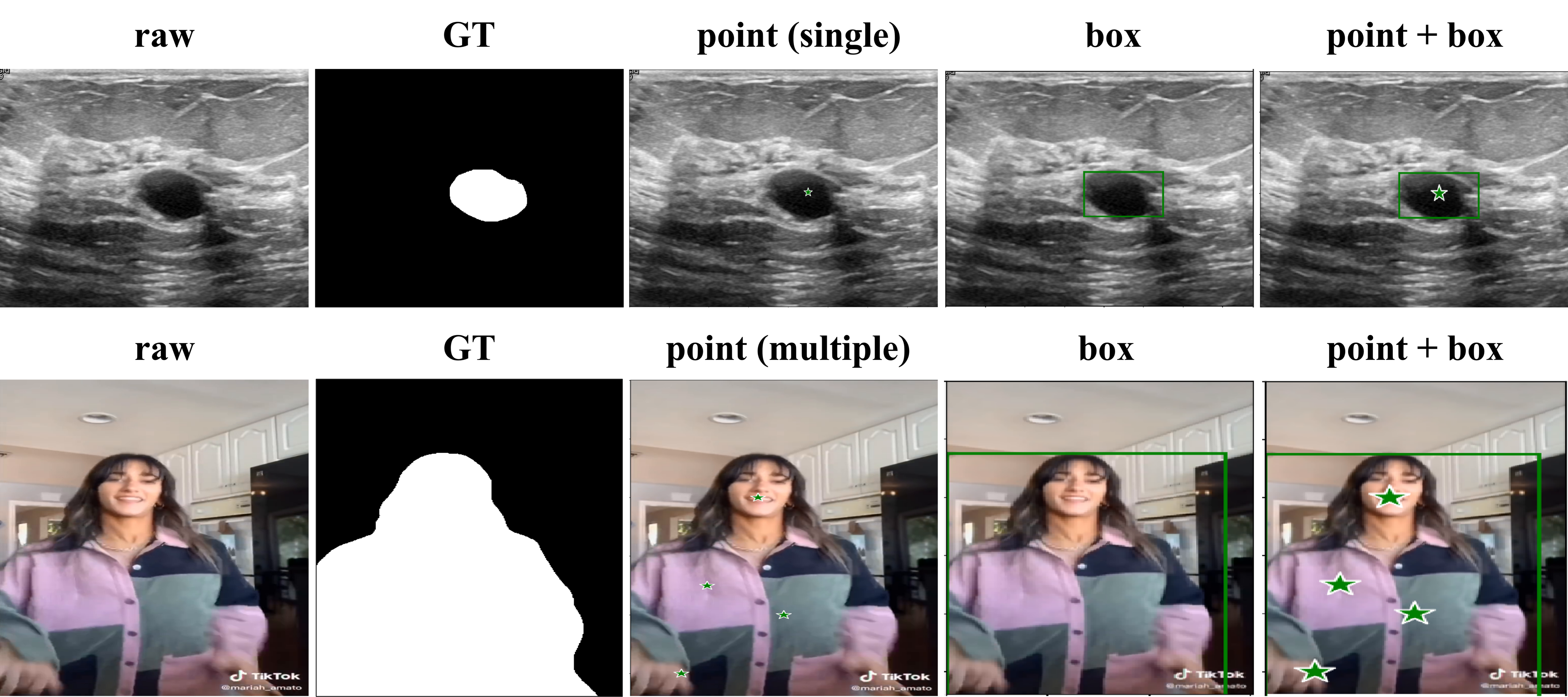}  
\caption{Examples of the three major types of prompting. Each row displays the raw image, ground truth mask, point, box, and combination of point and box prompting for a single example. For point prompting, we explore two variants: single-point prompting, as demonstrated in the first row, and multiple-point prompting, shown in the second row.}
\label{fig: prompting_images}
\end{figure*}

To select appropriate prompts, we carefully examine the ground truth mask and choose point locations or bounding boxes that best capture the object's features. Figure~\ref{fig: prompting_images} illustrates examples of the three major types of prompting, showing the raw image, ground truth mask, point, box, and combination of point and box prompting for two image examples.

\subsubsection{Image Perturbation Types}
We evaluate the robustness of SAM against various real-world challenges by considering a diverse set of image perturbations. These perturbations are categorized into six groups, encompassing a total of fifteen distinct types.

The Noise group includes Gaussian noise, which can appear in low-light conditions; shot noise, or Poisson noise, caused by the discrete nature of light; and salt-and-pepper noise, or impulse noise, resulting from bit errors. The Blur group comprises Gaussian blur, representing out-of-focus images; motion blur, occurring when the camera moves quickly; and defocus blur, which happens with unfocused lenses. Optical and Geometric (OG) perturbations include chromatic aberration, caused by lens dispersion; elastic transformations, which stretch or contract small image regions; and radial distortion, originating from the camera's lens geometry. Illumination and Color (IC) perturbations consist of variations in brightness and contrast due to lighting conditions and the photographed object's color, as well as changes in saturation. Environmental (ENV) perturbations encompass snow, a visually obstructive form of precipitation, and fog, which shrouds objects and affects visibility. Lastly, the Compression (CMP) group includes JPEG compression artifacts that result from lossy image compression.

\begin{figure*}[htbp]
\centering
\includegraphics[width=\textwidth]{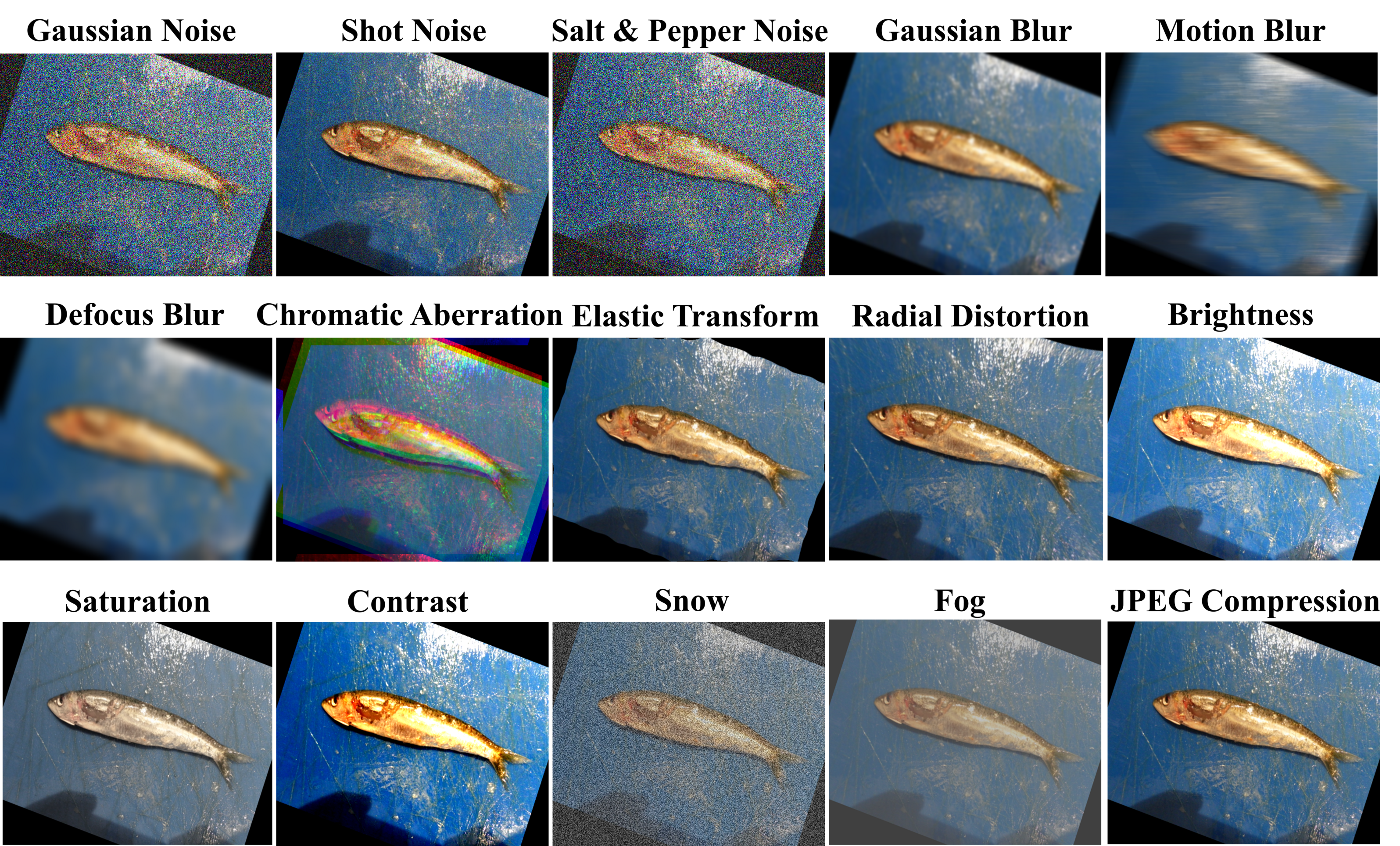}  
\caption{Illustration of fifteen perturbation types.}
\label{fig: perturbation_types}
\end{figure*}

Table \ref{tab: perturbation_types} provides an overview of the fifteen perturbation types and their corresponding parameters used in the experiments. In addition, Figure \ref{fig: perturbation_types} presents visual examples of each perturbation type, illustrating the effects of these perturbations on a sample image. This extensive selection of perturbations enables us to thoroughly assess SAM's robustness and adaptability in handling a wide range of real-world challenges.

\begin{table*}[htbp]
\centering
\caption{Overview of six categories of image perturbations, comprising a total of fifteen types and their respective parameters used in the experiments. Here, OG, IC, ENV, and CMP stand for Optical and Geometric, Illumination and Color, Environmental, and Compression, respectively.}
\resizebox{\linewidth}{!}{%
\label{tab: perturbation_types}
\begin{tabular}{cccc}
\hline
\textbf{Category} & \textbf{Perturbation} & \textbf{Abbreviation} & \textbf{Parameters} \\ \hline
\multirow{3}{*}{Noise} & Gaussian Noise & GN & mean=0, std.=100 \\ 
 & Shot Noise & SN & intensity=0.1 \\ 
 & Salt \& Pepper Noise & SPN & prob=0.15 \\ \hline
\multirow{3}{*}{Blur} & Gaussian Blur & GB & kernel size=15 \\
 & Motion Blur & MB & kernel size=45 \\ 
 & Defocus Blur & DB & kernel size=35, sigma\_x=sigma\_y=10 \\ \hline
\multirow{3}{*}{OG} & Chromatic Aberration & CA & shift\_x= shift\_y=15 \\ 
 & Elastic Transform & ET & alpha=100, sigma=10 \\ 
 & Radial Distortion & RD & k1=$-0.5$, k2=0.05, k3=p1=p2=0 \\ \hline
\multirow{3}{*}{IC} & Brightness & BRT & factor=1.5 \\ 
 & Saturation & SAT & coefficient=0.5 \\ 
 & Contrast & CON & factor=2 \\ \hline
\multirow{2}{*}{ENV} & Snow & SNOW & coefficient=0.5 \\ 
 & Fog & FOG & intensity=0.5 \\ \hline
CMP & JPEG Compression & COM & quality=15 \\ \hline
\end{tabular}}
\end{table*}

\subsection{Evaluation Metrics}
We evaluate the performance of SAM using several common metrics for image segmentation tasks, including Intersection over Union (IoU), recall, precision, and F1 score. Intersection over Union (IoU), also known as the Jaccard index, represents the ratio of the area of intersection to the area of union between the predicted segmentation mask and the ground truth mask. IoU is defined as:

\begin{align*}
    IoU = \frac{TP}{TP + FP + FN},
\end{align*}
where TP (true positive) denotes the number of correctly classified pixels, FP (false positive) represents the number of pixels incorrectly classified as positive, and FN (false negative) corresponds to the number of pixels incorrectly classified as negative.

Recall is the ratio of true positive pixels to the total number of actual positive pixels, measuring the model's ability to detect the object of interest. Recall is defined as:

\begin{align*}
    Recall = \frac{TP}{TP + FN}.
\end{align*}

Precision is the ratio of true positive pixels to the total number of pixels classified as positive, reflecting the model's ability to correctly identify the object of interest. Precision is defined as:

\begin{align*}
    Precision = \frac{TP}{TP + FP}.
\end{align*}

Finally, the F1 score, also known as the harmonic mean of precision and recall, combines both precision and recall into a single metric, providing a balanced view of the model's performance. The F1 score is defined as:

\begin{align*}
    F1 = 2 \times \frac{precision \times recall}{precision + recall}.
\end{align*}

\section{Results}
In this section, we present a comprehensive analysis of SAM's performance and robustness from various perspectives. First, we provide an overall performance comparison between raw and perturbed images to evaluate the impact of image perturbations on the model's segmentation capability. Next, we evaluate SAM's vulnerabilities and stability by examining its performance across different perturbation types. Then, we assess the effect of perturbation severity on model resilience, exploring how the model copes with varying degrees of distortion. Furthermore, we investigate the influence of different prompting techniques on model robustness, highlighting the importance of selecting the appropriate prompting strategy to enhance performance and mitigate perturbations. Lastly, we conduct a dataset-specific robustness analysis to uncover how SAM's performance varies across datasets with unique characteristics, emphasizing the need for tailored solutions.  

\subsection{Overall Performance Comparison: Raw vs. Perturbed Images}

We present an overall comparison of SAM's performance on raw and perturbed images across nine diverse datasets. Table~\ref{tab: overall_performance} displays the average performance of SAM on both raw and perturbed images across different datasets, taking into account IoU, F1, precision, and recall. For each dataset, we first compute the performance of the predicted maps using these metrics and then average the results. For perturbed image performance, we report the averaged performance across all perturbation settings. From the table, we observe that the performance across the majority of datasets and metrics is generally lower for perturbed images compared to raw images. This suggests that SAM tend to struggle when handling perturbed images, indicating potential vulnerabilities and a lack of robustness.

\begin{table*}[htbp]
\centering 
    \caption{Comparison of overall performance for raw images and perturbed images (Pert.) across various datasets, using IoU, F1, Recall, and Precision as evaluation metrics.}
  \resizebox{\linewidth}{!}{%
  \begin{tabular}{ccccccccc}
    \toprule
    \multirow{2}{*}{\textbf{Dataset}} & \multicolumn{2}{c}{\textbf{IoU}} & \multicolumn{2}{c}{\textbf{ F1}} & \multicolumn{2}{c}{\textbf{Recall}} & \multicolumn{2}{c}{\textbf{Precision}} \\
\cmidrule(lr){2-3}\cmidrule(lr){4-5}\cmidrule(lr){6-7} \cmidrule(lr){8-9}  & raw & Pert. & raw & Pert. & raw & Pert. & raw & Pert.  \\ \midrule Forest Aerial & 0.652 & 0.644 & 0.729 & 0.725 & 0.747 & 0.822 & 0.792 & 0.751\\ Water Bodies & 0.472 & 0.456 & 0.586 & 0.571 & 0.679 & 0.671 & 0.574 & 0.570 \\ Road Extraction & 0.126 & 0.090 & 0.192 & 0.148 & 0.688 & 0.769 & 0.129 & 0.091 \\ Breast Ultrasound & 0.665 & 0.644 & 0.774 & 0.757 & 0.735 & 0.740 & 0.903 & 0.869 \\ Chest X-Ray & 0.735 & 0.574 & 0.834 & 0.692 & 0.868 & 0.715 & 0.856 & 0.743 \\ Fish & 0.908 & 0.892 & 0.940 & 0.936 & 0.924 & 0.930 & 0.983 & 0.959 \\ Fire & 0.165 & 0.153 & 0.266 & 0.247 & 0.571 & 0.592 & 0.239 & 0.216 \\ Crack & 0.109 & 0.111 & 0.163 & 0.164 & 0.714 & 0.801 & 0.117 & 0.119 \\ TikTok Dancing & 0.927 & 0.905 & 0.962 & 0.949 & 0.930 & 0.920 & 0.997 & 0.983 \\
  \bottomrule
\end{tabular}}
\label{tab: overall_performance}
\end{table*}

The decline in performance is more prominent in certain datasets, such as the Chest X-Ray dataset, which shows a considerable drop in IoU, F1, recall, and precision. These findings emphasize the importance of investigating the underlying factors contributing to the performance degradation and seeking specific strategies to improve model robustness against perturbations, especially for critical applications like medical imaging.

\begin{figure*}[htbp]
\centering
\includegraphics[width=\textwidth]{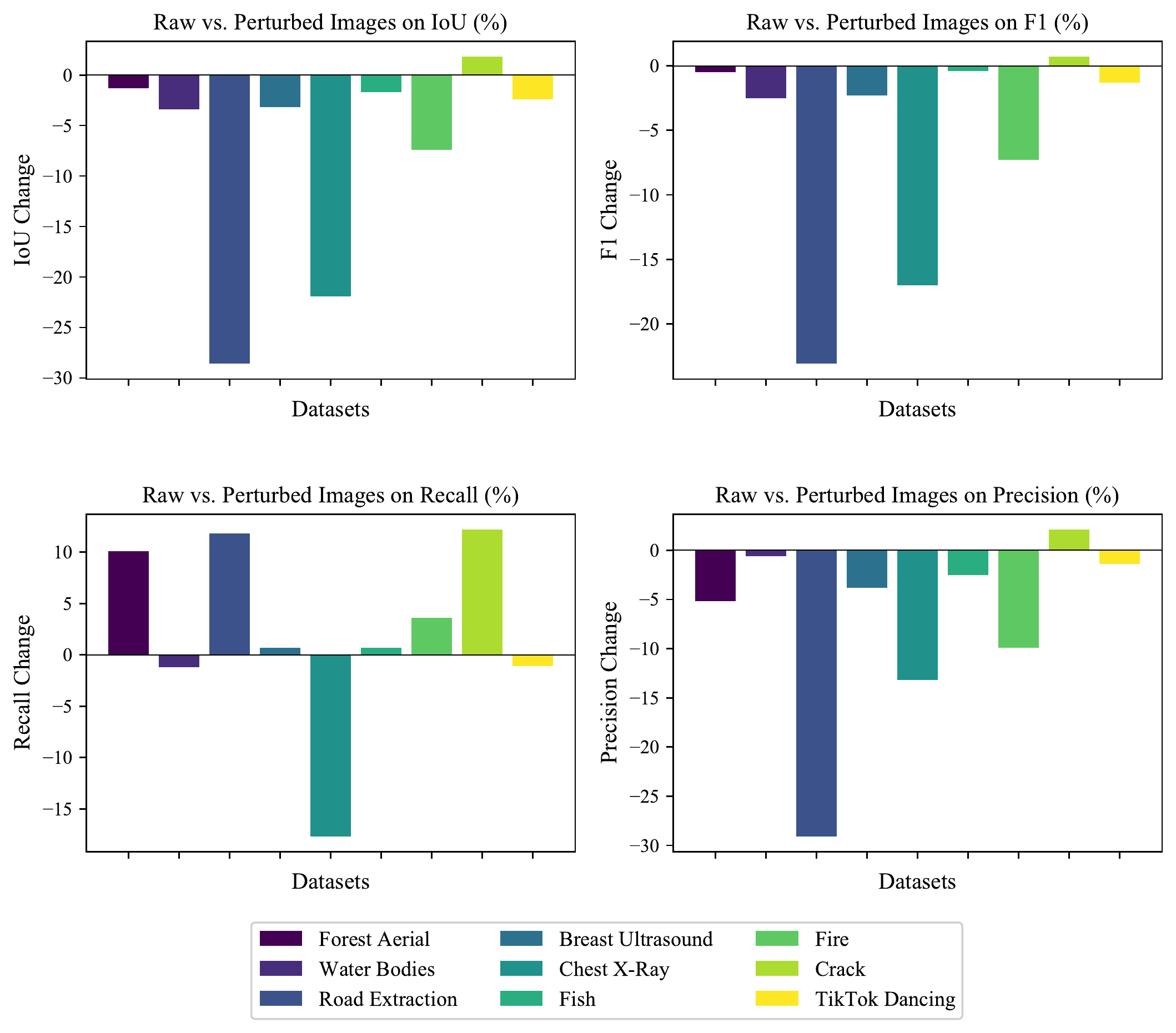} 
\caption{Percentage change in performance metrics (IoU, F1, Recall, and Precision) between raw and perturbed images across the nine datasets. The figure illustrates that performance typically declines for perturbed images, with some exceptions where it slightly increases.}
\label{fig: performance_change}
\end{figure*}

Furthermore, Figure \ref{fig: performance_change} offers a detailed perspective on the percentage changes in performance metrics between raw and perturbed images across the nine datasets, further illustrating the impact of perturbations on model robustness. We observe a general trend of decreasing performance with the introduction of perturbations, with some notable exceptions. For instance, the Chest X-Ray dataset experiences a drop of 21.9\% in IoU and 17.0\% in F1 score, indicating a significant compromise in the model's robustness. However, certain datasets, such as Crack, display a relative resilience to perturbations, with a marginal increase in mean IoU by 1.8\% and F1 score by 0.7\%.

These initial observations lay the foundation for the remainder of this section, where we delve deeper into various aspects of model vulnerabilities, stability, and resilience. We also explore the impact of prompting techniques and dataset-specific factors on model robustness, providing a more holistic understanding of the challenges and potential avenues for improvement in this domain.

\subsection{Model Vulnerabilities and Stability to Perturbations}
We investigate the vulnerability and stability of the SAM model under various perturbations by examining its performance metrics and qualitative examples. Table~\ref{tab: change_per_perturbation} presents the average percentage changes in IoU, F1, recall, and precision for each perturbation type compared to raw images, averaged across all datasets. 

\begin{table*}[htbp]
\centering
\caption{Average percentage changes in IoU, F1, Recall, and Precision for each perturbation type compared to raw images, averaged over all datasets. Bold numbers indicate the top-3 most negative percentage changes, highlighting SAM's most vulnerable perturbations. Underlined numbers denote the bottom-3 least negative or positive percentage changes, representing the least vulnerable perturbations. SAM is highly susceptible to chromatic aberration, motion blur, and Gaussian noise perturbations, while showing increased robustness against brightness and saturation change perturbations.}
\resizebox{\linewidth}{!}{%
\begin{tabular}{ccccc}
\toprule
\textbf{Perturbation} & \textbf{IoU (\%)} & \textbf{F1 (\%)} & \textbf{Recall (\%)} & \textbf{Precision (\%)} \\
\midrule
Brightness & \underline{0.4} & \underline{0.2} & $\mathbf{-1.8}$ & 1.9 \\
Chromatic aberration & $\mathbf{-14.5}$ & $\mathbf{-11.0}$ & 2.7 & $-12.4$  \\
JPEG Compression & $-5.0$ & $-4.3$ & 0.6 & \underline{$-1.1$} \\
Contrast & $-0.9$ & \underline{$-0.5$} & $\mathbf{-1.6}$ & \underline{0.8} \\
Defocus blur & $-13.4$ & $-10.9$ & \underline{7.2} & $-13.3$\\
Motion blur & $\mathbf{-17.4}$ & $\mathbf{-13.2}$ & \underline{9.0} & $\mathbf{-19.2}$\\
Gaussian noise & $\mathbf{-15.6}$ & $\mathbf{-12.5}$ & \underline{4.4} & $\mathbf{-15.3}$  \\
Salt pepper noise & $-13.9$ & $-10.8$ & 4.1 & $\mathbf{-13.7}$ \\
Elastic transform & $-2.8$ & $-2.2$ & 1.6 & $-1.8$\\
Radial distortion & $-12.6$ & $-9.7$ & $\mathbf{-1.8}$ & $-13.5$   \\
Gaussian blur & \underline{$0.1$} & \underline{$0.0$} & 2.4 & $-0.9$ \\
Saturation  & \underline{$-0.5$} & \underline{$-0.5$} & $-0.1$ & \underline{$0.0$}        \\
Shot noise & $-5.1$ & $-3.7$ & 0.5 & $-3.3$ \\
Snow & $-11.6$ & $-9.4$ & 2.8 & $-12.3$\\
Fog & $-0.7$ & $-0.9$ & 1.8 & $-1.8$\\
\bottomrule
\end{tabular}}
\label{tab: change_per_perturbation}
\end{table*}

Our findings reveal that SAM experiences the most substantial performance decline in terms of IoU, F1, and Precision when exposed to chromatic aberration, motion blur, and Gaussian noise perturbations. This observation suggests that SAM's capability to cope with these particular distortions or noise is not optimal, necessitating further exploration into the root causes and potential remedies. This susceptibility might stem from the intricate distortions and noise introduced by these perturbations, which can substantially challenge the model's capacities, hindering its consistent performance under such conditions.

In contrast, SAM exhibits greater robustness when exposed to brightness and saturation change perturbations, displaying minimal negative or even positive percentage changes in the metrics. This observation implies that the model's performance remains relatively stable under these conditions. Investigating the factors contributing to SAM's robustness in these scenarios is crucial to uncover potential approaches for enhancing its performance under more demanding perturbations.

Moreover, Table \ref{tab: cv_per_perturbation} complements our understanding of the model's stability by examining the coefficient of variation (CV) for each perturbation for each metric, averaged over all datasets. The CV represents the ratio of the standard deviation to the mean. Higher CV values for Gaussian noise, defocus blur, and motion blur perturbations indicate greater variability in SAM's performance across different datasets, highlighting its vulnerability under these conditions. Conversely, the consistently lower CV values for brightness, Gaussian blur, and contrast reveal relative stability in SAM's performance, suggesting that it is less affected by these perturbations.

\begin{figure*}[htbp]
\centering
\includegraphics[width=\textwidth]{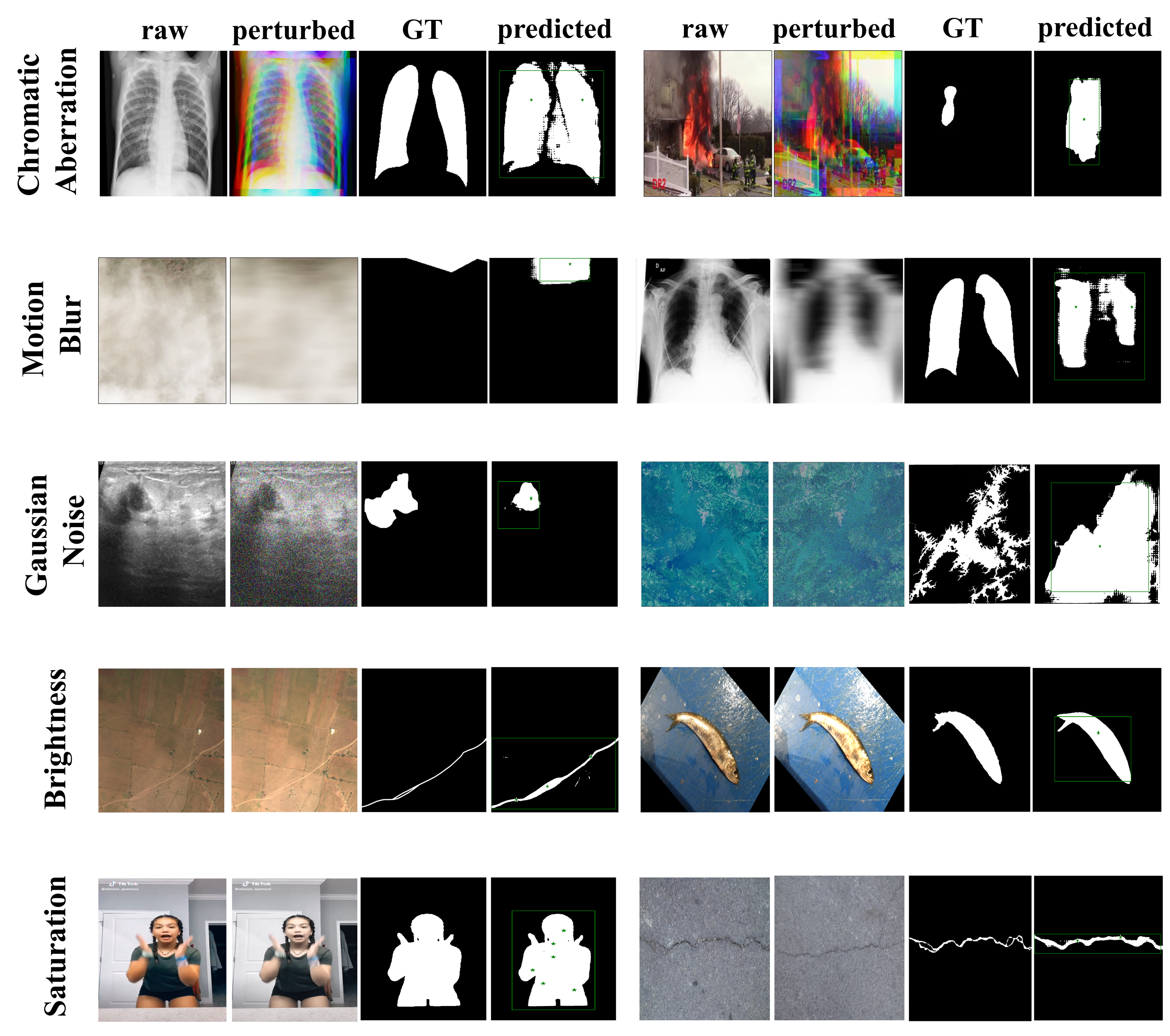}  
\caption{Visual case examples of SAM's segmentation performance under various perturbations. Each row represents a perturbation type, with the first three rows illustrating challenging cases (chromatic aberration, motion blur, and Gaussian noise), and the last two rows showing robust cases (brightness and saturation change). Two examples from different datasets are provided per row. Each example comprises four components: the raw image, perturbed image, ground truth mask, and predicted mask under perturbation.}
\label{fig: model_vulnerability_example}
\end{figure*}

The visual case examples shown in Figure \ref{fig: model_vulnerability_example} provide further insights into SAM's segmentation performance under various perturbations. The challenging cases involving chromatic aberration, motion blur, and Gaussian noise, as illustrated in the first three rows, show that the model struggles to produce accurate segmentation masks. These cases reveal that the perturbations disrupt the model's ability to recognize and segment the objects of interest, pointing to the need for additional research to enhance its resilience. On the other hand, the last two rows display more robust cases involving brightness and saturation change perturbations, where the model's performance remains relatively unaffected.

Our analysis of SAM's vulnerabilities and stability to perturbations reveals key areas for improvement and potential directions for enhancing the model's robustness. Future research should prioritize addressing the most challenging perturbations, such as chromatic aberration, motion blur, and Gaussian noise, and focus on understanding the factors that contribute to the model's stability under specific perturbations. This will enable the development of more robust and reliable segmentation models that can better handle a wide range of real-world conditions.

\begin{table*}[htbp]
\centering
\caption{Coefficient of variation (CV) percentages for IoU, F1, Recall, and Precision are presented for each perturbation, averaged over all datasets. Bold numbers indicate the top-3 highest CV values, representing greater instability in SAM's performance, while underlined numbers denote the bottom-3 lowest CV values, suggesting more stable model performance for each metric. Higher CV values for Gaussian noise, defocus blur, and motion blur perturbations reveal model vulnerabilities, while consistently lower CV values for brightness, Gaussian blur, and contrast demonstrate relative model stability.}
\resizebox{\linewidth}{!}{%
\begin{tabular}{ccccc}
\toprule
\textbf{Perturbation} & \textbf{IoU (CV\%)} & \textbf{F1 (CV\%)} & \textbf{Recall (CV\%)} & \textbf{Precision (CV\%)} \\
\midrule
Brightness & \underline{57.9} & \underline{49.6} & 15.1 & \underline{55.4} \\
Chromatic aberration & 60.3 & 53.0 & 15.3 & 57.3\\
JPEG Compression & 61.0 & 52.2 & 15.9 & 56.4 \\
Contrast & \underline{58.0} & \underline{49.4} & 14.9 & \underline{55.3} \\
Defocus blur & $\mathbf{65.1}$ & 54.5 & $\mathbf{18.9}$ & $\mathbf{60.9}$\\
Motion blur & $\mathbf{64.6}$ & $\mathbf{54.9}$ & 15.5 & 62.4\\
Gaussian noise & 64.0 & $\mathbf{55.0}$ & $\mathbf{18.2}$ & $\mathbf{60.8}$ \\
Salt pepper noise & $\mathbf{66.2}$ & $\mathbf{55.1}$ & 17.0 & $\mathbf{61.1}$ \\
Elastic transform & 58.8 & 50.4 & \underline{14.2} & 56.1\\
Radial distortion & 61.4 & 53.6 & $\mathbf{18.1}$ & 58.4 \\
Gaussian blur & \underline{58.4} & \underline{49.9} & \underline{14.3} & 56.2 \\
Saturation  & 58.6 & 50.3 & 15.1 & \underline{55.8} \\
Shot noise & 60.7 & 51.8 & 16.5 & 57.1 \\
Snow & 64.5 & 54.8 & 17.3 & 60.6\\
Fog & 58.5 & 50.4 & \underline{14.3} & 56.2 \\
\bottomrule
\end{tabular}}
\label{tab: cv_per_perturbation}
\end{table*}

\subsection{Effect of Perturbation Severity on Model Resilience}
We assess the impact of perturbation severity on the resilience of SAM by selecting five perturbation types, each representing a different category of image distortions commonly encountered in real-world scenarios. The chosen perturbations include Elastic Transformation (geometric distortion), Chromatic Aberration (optical distortion), Motion Blur (motion-based distortion), Gaussian Noise (noise-based distortion), and Brightness Change (illumination-based distortion). For each perturbation, we consider three severity levels: low, medium, and high. Table~\ref{tab: severity_parameter} presents the perturbation types and their corresponding parameters selected for the experiments. The experiments are conducted on four datasets: Forest Aerial, Fire, Chest X-Ray, and TikTok Dancing. We select these datasets to cover a wide range of image characteristics, such as simple single-object scenes, complex multi-object compositions, and cluttered backgrounds.

\begin{figure*}[htbp]
\centering
\includegraphics[width=\textwidth]{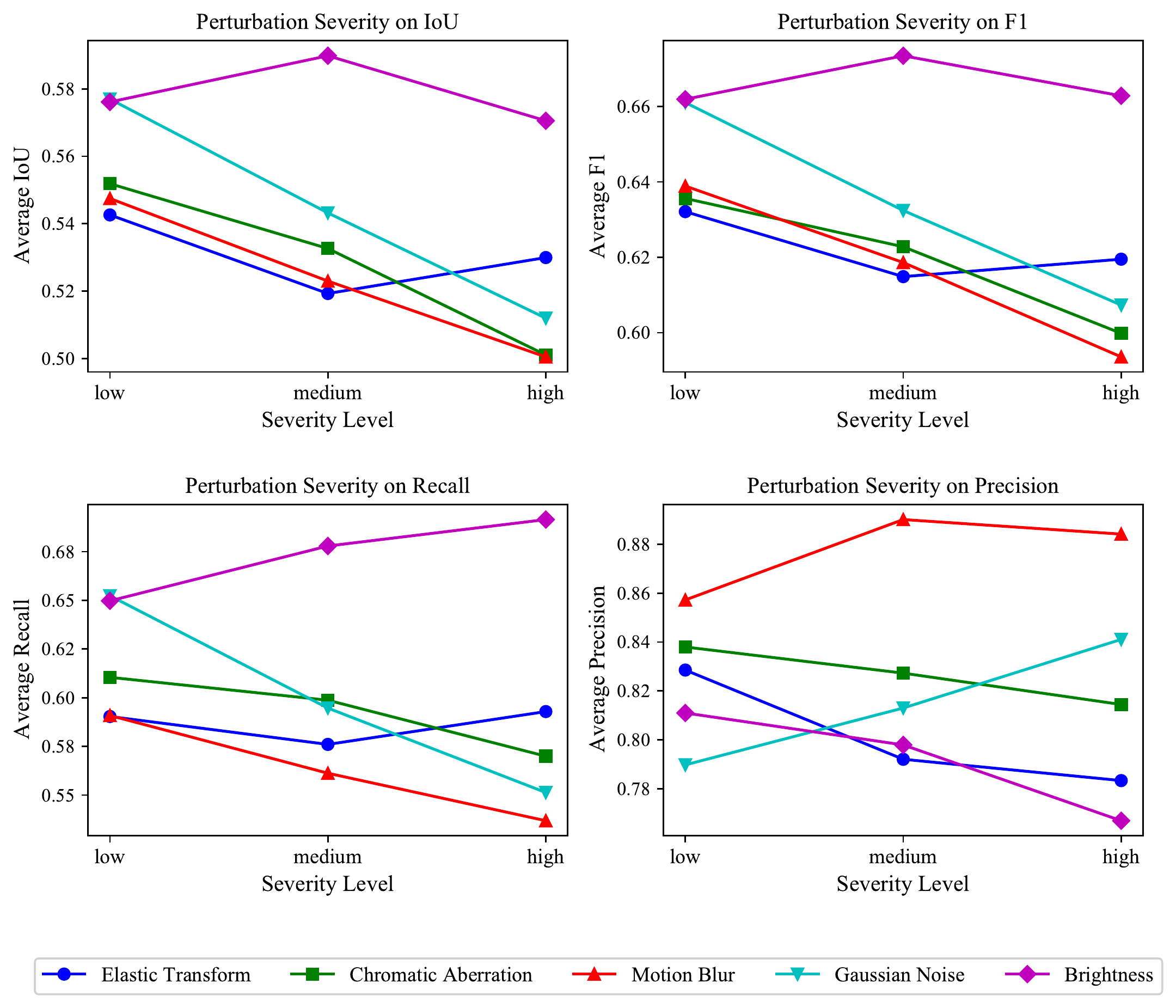}  
\caption{Impact of perturbation severity on IoU, F1, Recall, and Precision for five perturbation types (Elastic Transform, Chromatic Aberration, Motion Blur, Gaussian Noise, and Brightness), averaged across four datasets (Forest Aerial, Fire, Chest X-Ray, and TikTok Dancing). Although the performance of each metric generally decreases with increasing perturbation severity, some cases exhibit better performance at higher intensity levels.}
\label{fig: perturbation_severity}
\end{figure*}

\begin{table*}[htbp]
\centering
\caption{Perturbation types with associated parameters for low, medium, and high severity levels used in experiments.}
\resizebox{\linewidth}{!}{%
\begin{tabular}{ccccc}
\hline
\textbf{Perturbation} & \textbf{Parameters} & \textbf{Low} & \textbf{Medium} & \textbf{High} \\
\hline
Chromatic Aberration & (shift\_x, shift\_y) & (5, 5) & (10, 10) & (15, 15) \\
Motion Blur & kernel size & 15 & 30 & 45 \\
Gaussian Noise & (mean, std.) & (0, 10) & (0, 50) & (0, 100)\\
Elastic Transform & (alpha, sigma) & (25, 2) & (100, 4) & (200, 6) \\
Brightness & factor & 1.2 & 1.5 & 2.0 \\
\hline
\end{tabular}}
\label{tab: severity_parameter}
\end{table*}

Figure~\ref{fig: perturbation_severity} demonstrates the varying impact of perturbation severity on the performance of SAM across four metrics (IoU, F1, recall, and precision) for five types of perturbations.
Averaged over four datasets, the results generally indicate a decline in performance as the severity of perturbations increases. Interestingly, some cases exhibit better performance at higher intensity levels, which could be attributed to a variety of factors. For example, the brightness perturbation shows an improvement in F1 scores from 0.662 at low severity to 0.674 at medium severity. The recall score for brightness perturbation also increases consistently across severity levels. This unexpected improvement might be due to the SAM's ability to adapt to certain types of illumination changes or the presence of inherent robustness in SAM for specific perturbations. In contrast, metrics like IoU and precision for chromatic aberration and motion blur display more consistent declines as the severity levels increase. These trends suggest that the SAM's resilience can be influenced by the nature and severity of the perturbations. It also highlights the importance of considering diverse image distortions and understanding their impact on SAM's performance.

The observed improvements in performance at higher intensity levels for certain cases suggest that SAM's resilience is not uniform across all types of perturbations. The presence of underlying SAM's robustness or adaptability to specific perturbations may be responsible for these performance improvements. Consequently, a one-size-fits-all approach is insufficient for evaluating SAM. A comprehensive assessment using diverse perturbation types and severity levels is essential for understanding SAM's performance and adaptability in real-world situations.

\subsection{Influence of Prompting Techniques on Model Robustness}
Renowned for its promptable segmentation capabilities, we investigate the performance of the SAM model using three major prompting methods: Point, Box, and a Combination of Point and Box. This investigation aims to understand the role of prompting techniques in enhancing the model's robustness and overall performance.

Table~\ref{tab: prompting} presents the performance of the SAM model using three major prompting techniques on raw and perturbed images, along with the percentage change in performance on IoU, F1, recall, and precision. The results reveal that the Combination of Point and Box prompting method consistently yields superior results across all performance metrics for both types of images. Furthermore, this prompting technique demonstrates increased resilience to perturbations, as indicated by the smaller percentage drop in performance compared to the other techniques.

\begin{table*}[htbp]
\centering
\caption{Comparison of SAM model performance with three prompting techniques (Point, Box, and Combination of Point and Box) on raw and perturbed images, including the percentage change in performance. The Combination of Point and Box technique demonstrates better overall performance and improved robustness against perturbations.}
\begin{tabular}{ccccc}
\toprule
\textbf{Prompting Technique} &    \textbf{Metric} &   \textbf{Raw} &  \textbf{Pert.} &  \textbf{Change (\%)} \\
\midrule \multirow{4}{*}{Point} & IoU & 0.497 &      0.462 & $-7.1$ \\ & F1 & 0.574 & 0.540 & $-6.0$ \\ & Recall & 0.753 & 0.766 & 1.7 \\ & Precision & 0.610 & 0.579 & $-5.0$ \\ \midrule \multirow{4}{*}{Box} & IoU & 0.536 &      0.492 & $-8.1$ \\ & F1 & 0.610 & 0.572 & $-6.4$ \\ &    Recall & 0.740 & 0.745 & 0.7 \\ & Precision & 0.622 &      0.574 & $-7.7$ \\ \midrule \multirow{4}{*}{\makecell{Combination of \\ Point and Box}} & IoU & 0.554 & 0.522 & $-5.7$ \\ &        F1 & 0.631 & 0.606 & $-4.0$ \\ &    Recall & 0.791 &      0.814 & 2.8 \\ & Precision & 0.632 & 0.598 & $-5.3$ \\    
\bottomrule
\end{tabular}
\label{tab: prompting}
\end{table*}

\begin{figure*}[htbp]
\centering
\includegraphics[width=\textwidth]{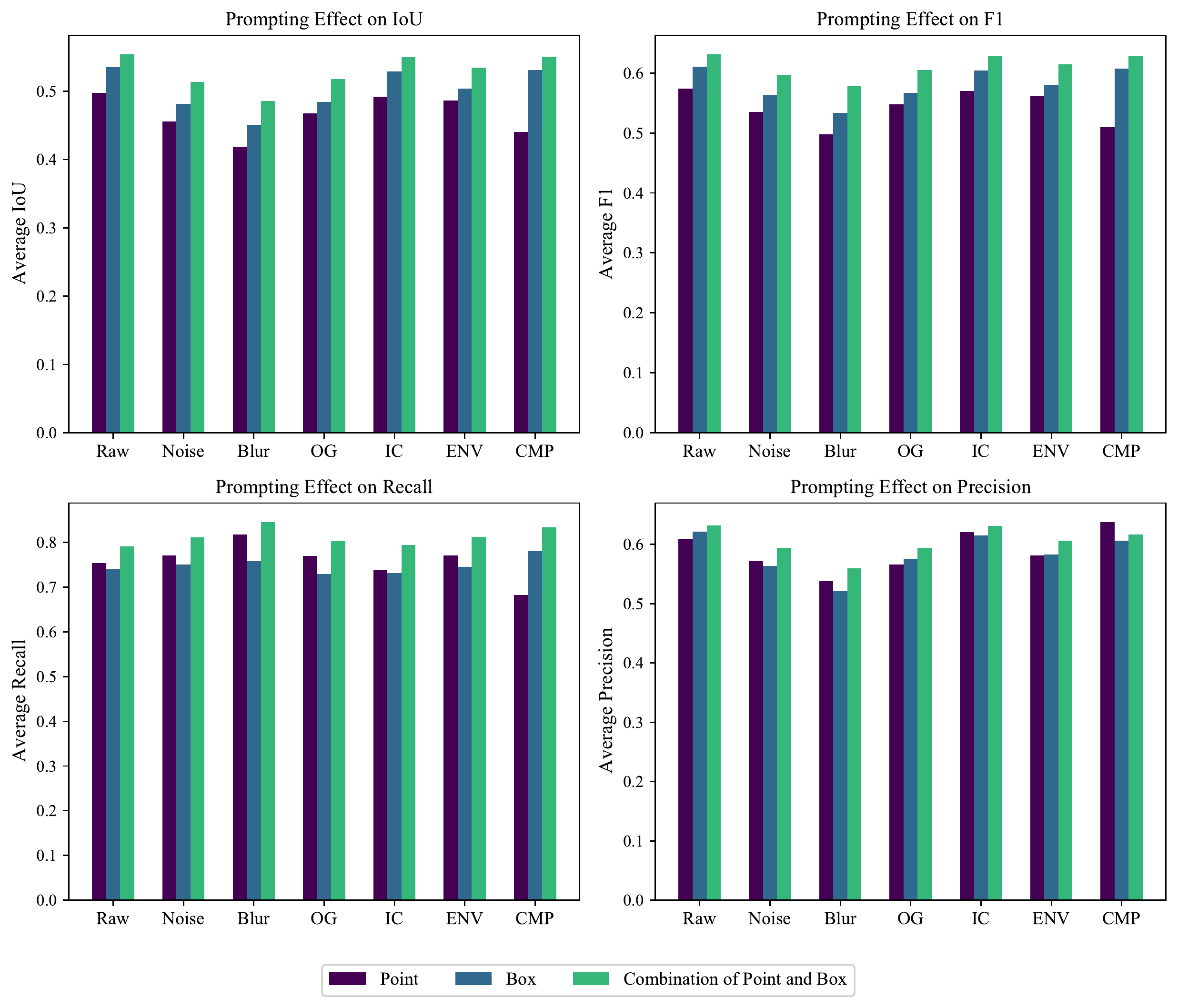}  
\caption{The impact of different prompting techniques (Point, Box, and Combination of Point and Box) on model performance across various perturbation categories, including Noise, Blur, Optical and Geometric (OG), Illumination and Color (IC), Environmental (ENV), and Compression (CMP). The Combination of Point and Box prompting consistently outperforms the other techniques across all categories, highlighting its effectiveness and robustness in handling a diverse set of image perturbations.}
\label{fig: prompting_effect}
\end{figure*}

A more detailed analysis is presented in Figure~\ref{fig: prompting_effect}, where we explore the impact of different prompting techniques on SAM's performance across various perturbation categories, including Noise, Blur, Optical and Geometric, Illumination and Color, Environmental, and Compression. We observe that the combination of Point and Box prompting consistently outperforms other methods across all perturbation categories, demonstrating its effectiveness and robustness. This finding suggests that the combined prompting approach is better suited to manage the diverse challenges posed by different perturbation types, thereby enhancing the overall robustness of SAM.

Our results highlight the role that prompting plays in augmenting the SAM model's promptable segmentation capabilities. By integrating Point and Box prompting techniques, we can achieve a certain extent of improvement in both model performance and robustness. This underscores the value of selecting and designing appropriate prompting strategies to ensure that a model can effectively handle a diverse array of image perturbations while maintaining high performance.

\subsection{Dataset-Specific Robustness Analysis}

Due to the unique characteristics of each dataset, we study the dataset-specific impact on SAM's vulnerability and robustness to various perturbations. Table~\ref{tab: data_specific} illustrates the preferred prompt type, top-3 vulnerable perturbations, and top-3 robust perturbations for each dataset. From the table, our aim is to uncover the intrinsic features that make a dataset more or less susceptible to specific perturbations. In addition, we observe that a majority of the datasets achieve better performance with a combination of box and point prompts, while Forest Aerial, Fire, and Fish datasets show a preference for box prompts. This finding suggests the importance of tailoring prompting strategies according to the unique characteristics of each dataset.

Forest, Fire, and Crack datasets are influenced by texture and color. Forest images with complex patterns and high variation are susceptible to motion blur and elastic transform. The high contrast and irregular shapes of fire images make them prone to chromatic aberration and motion blur but more robust against elastic transform and Gaussian blur. Crack images, with linear patterns and grayscale colors, are susceptible to Gaussian noise and elastic transform but robust against contrast, shot noise, and chromatic aberration.

Road and Chest X-Ray datasets exhibit similarities in their vulnerability to perturbations. Road images, with smooth surfaces and grayscale colors, are susceptible to radial distortion, Gaussian noise, and salt pepper noise but robust against saturation, fog, and Gaussian blur. Radiology images, with smooth gradients and subtle patterns, are vulnerable to salt pepper noise, defocus blur, and motion blur but robust against fog, saturation, and compression.

The Breast Ultrasound dataset stands out due to its unique speckle noise and low contrast, making it susceptible to radial distortion and motion blur. The grayscale colors contribute to the vulnerability to defocus blur, but the presence of organ shapes and structures make the dataset more robust against snow, shot noise, and brightness perturbations.

Fish and TikTok Dancing datasets exhibit diverse colors and backgrounds, making them susceptible to chromatic aberration, radial distortion, and motion blur. Fish images' varying shapes and patterns make the dataset more robust against compression, fog, and brightness, while Dancing images' diverse stage elements make the dataset more robust against elastic transform, compression, and Gaussian blur.

Lastly, the Water bodies dataset, with its wavy patterns and reflections, is more susceptible to chromatic aberration and motion blur, as these perturbations can further distort the existing patterns. Additionally, the shades of blue and green in the images make them vulnerable to Gaussian noise, which can compromise the subtle color differences. However, the presence of aquatic life or vegetation adds complexity to the images, making them more robust against perturbations such as snow, shot noise, and contrast, as these elements can still be distinguishable despite the perturbations.

Our findings emphasize the need to take into account the unique characteristics of each dataset when analyzing SAM's performance. Gaining insights into the vulnerabilities and robustness of each dataset under specific perturbations can assist in developing more robust models that address the distinct challenges posed by different datasets. Utilizing domain knowledge tailored to each particular dataset can improve our comprehension of these vulnerabilities and contribute to designing more effective solutions that account for real-world complexities and variations.

\begin{table*}[htbp]
\centering
\caption{Preferred prompt type, Top-3 vulnerable perturbations (Pert.), and Top-3 robust perturbations for each dataset, determined by majority voting among four metrics (IoU, F1-score, precision, and recall). Results reveal varying vulnerable and robust perturbations across datasets. Across all datasets, motion blur (MB) and chromatic aberration (CA) frequently emerge as vulnerable perturbations, while shot noise (SN) and compression (COM) commonly appear as robust perturbations.}
\resizebox{\linewidth}{!}{%
\begin{tabular}{cccc}
\hline
\textbf{Dataset} & \textbf{Preferred Prompt} & \textbf{Top-3 Vulnerable Pert.} & \textbf{Top-3 Robust Pert.} \\ \hline
Forest Aerial & Box & ET, MB, DB & SN, COM, GB \\
Water Bodies & Point $+$ Box & CA, MB, GN & SN, SNOW, CON\\
Road Extraction & Point $+$ Box & RD, GN, SPN & SAT, FOG, GB \\ 
Breast Ultrasound & Point $+$ Box & RD, MB, DB & SNOW, SN, BRT \\
Chest X-Ray & Point $+$ Box & SPN, DB, MB & FOG, SAT, COM\\
Fish & Box & CA, RD, MB & COM, FOG, BRT \\
Fire & Box & CA, SPN, MB & ET, GB, CON \\ 
Crack & Point $+$ Box & GN, ET, MB & CON, SN, CA \\
TikTok Dancing & Point $+$ Box & CA, MB, RD & ET, COM, GB \\
\hline
\end{tabular}}
\label{tab: data_specific}
\end{table*}

\section{Discussion}
In this study, we investigated the robustness of the SAM model under fifteen perturbations and nine datasets, focusing on the relationship between perturbations, model vulnerabilities, stability, and the influence of prompting techniques on robustness.

We find that SAM's performance generally declines under perturbed images compared to raw images, with the extent of degradation varying across different perturbations. This observation highlights the model's vulnerability to certain perturbations and the need to develop strategies to enhance its robustness. As perturbation severity increases, performance generally declines, with a few exceptions showing better performance at higher intensity.

Moreover, our investigation into different prompting techniques for SAM reveals the importance of customization. Most datasets achieve better performance with a combination of box and point prompts, while others benefit from alternative prompting strategies. These findings suggest that incorporating human-in-the-loop interactions and adaptive prompting through multiple iterations can improve model performance and robustness against perturbations.

Our analysis also emphasizes the significance of considering each dataset's unique characteristics, such as color, texture, and patterns. Different datasets exhibit varying vulnerabilities and robustness under specific perturbations, underscoring the need for tailored models addressing the specific challenges presented by different datasets. Incorporating domain knowledge into the model training process can further enhance our understanding of these vulnerabilities and aid in designing more effective solutions.

In future work, we aim to explore various ways to improve SAM's robustness. These include researching more adaptive prompting strategies, integrating expert feedback through human-in-the-loop interactions, and creating dataset-specific data augmentation techniques that mimic real-world perturbations. We also plan to incorporate knowledge about the applied perturbations during model training and use transfer learning to enhance the model's resilience to various perturbations and address domain-specific challenges more effectively. Our primary goal is to refine SAM to increase its robustness, enabling it to better manage a variety of datasets and maintain strong performance across numerous real-world situations.

%% The Appendices part is started with the command \appendix;
%% appendix sections are then done as normal sections
%% \appendix

%% \section{}
%% \label{}

%% If you have bibdatabase file and want bibtex to generate the
%% bibitems, please use
%%
\bibliographystyle{elsarticle-num} 
\bibliography{ref}

\begin{thebibliography}{10}
\expandafter\ifx\csname url\endcsname\relax
  \def\url#1{\texttt{#1}}\fi
\expandafter\ifx\csname urlprefix\endcsname\relax\def\urlprefix{URL }\fi
\expandafter\ifx\csname href\endcsname\relax
  \def\href#1#2{#2} \def\path#1{#1}\fi

\bibitem{bommasani2021opportunities}
R.~Bommasani, D.~A. Hudson, E.~Adeli, R.~Altman, S.~Arora, S.~von Arx, M.~S.
  Bernstein, J.~Bohg, A.~Bosselut, E.~Brunskill, et~al., On the opportunities
  and risks of foundation models, arXiv preprint arXiv:2108.07258 (2021).

\bibitem{brown2020language}
T.~Brown, B.~Mann, N.~Ryder, M.~Subbiah, J.~D. Kaplan, P.~Dhariwal,
  A.~Neelakantan, P.~Shyam, G.~Sastry, A.~Askell, et~al., Language models are
  few-shot learners, Advances in neural information processing systems 33
  (2020) 1877--1901.

\bibitem{openai2023gpt4}
OpenAI, Gpt-4 technical report (2023).
\newblock \href {http://arxiv.org/abs/2303.08774} {\path{arXiv:2303.08774}}.

\bibitem{radford2021learning}
A.~Radford, J.~W. Kim, C.~Hallacy, A.~Ramesh, G.~Goh, S.~Agarwal, G.~Sastry,
  A.~Askell, P.~Mishkin, J.~Clark, et~al., Learning transferable visual models
  from natural language supervision, in: International conference on machine
  learning, PMLR, 2021, pp. 8748--8763.

\bibitem{kirillov2023segment}
A.~Kirillov, E.~Mintun, N.~Ravi, H.~Mao, C.~Rolland, L.~Gustafson, T.~Xiao,
  S.~Whitehead, A.~C. Berg, W.-Y. Lo, et~al., Segment anything, arXiv preprint
  arXiv:2304.02643 (2023).

\bibitem{ma2023segment}
J.~Ma, B.~Wang, Segment anything in medical images, arXiv preprint
  arXiv:2304.12306 (2023).

\bibitem{ji2023sam}
G.-P. Ji, D.-P. Fan, P.~Xu, M.-M. Cheng, B.~Zhou, L.~Van~Gool, Sam struggles in
  concealed scenes--empirical study on" segment anything", arXiv preprint
  arXiv:2304.06022 (2023).

\bibitem{zhou2023can}
T.~Zhou, Y.~Zhang, Y.~Zhou, Y.~Wu, C.~Gong, Can sam segment polyps?, arXiv
  preprint arXiv:2304.07583 (2023).

\bibitem{deng2023segment}
R.~Deng, C.~Cui, Q.~Liu, T.~Yao, L.~W. Remedios, S.~Bao, B.~A. Landman, L.~E.
  Wheless, L.~A. Coburn, K.~T. Wilson, et~al., Segment anything model (sam) for
  digital pathology: Assess zero-shot segmentation on whole slide imaging,
  arXiv preprint arXiv:2304.04155 (2023).

\bibitem{kamann2020benchmarking}
C.~Kamann, C.~Rother, Benchmarking the robustness of semantic segmentation
  models, in: Proceedings of the IEEE/CVF conference on computer vision and
  pattern recognition, 2020, pp. 8828--8838.

\bibitem{zhang2016instance}
Z.~Zhang, S.~Fidler, R.~Urtasun, Instance-level segmentation for autonomous
  driving with deep densely connected mrfs, in: Proceedings of the IEEE
  Conference on Computer Vision and Pattern Recognition, 2016, pp. 669--677.

\bibitem{malhotra2022deep}
P.~Malhotra, S.~Gupta, D.~Koundal, A.~Zaguia, W.~Enbeyle, Deep neural networks
  for medical image segmentation, Journal of Healthcare Engineering 2022
  (2022).

\bibitem{wang2018deepigeos}
G.~Wang, M.~A. Zuluaga, W.~Li, R.~Pratt, P.~A. Patel, M.~Aertsen, T.~Doel,
  A.~L. David, J.~Deprest, S.~Ourselin, et~al., Deepigeos: a deep interactive
  geodesic framework for medical image segmentation, IEEE transactions on
  pattern analysis and machine intelligence 41~(7) (2018) 1559--1572.

\bibitem{cao2020ship}
X.~Cao, S.~Gao, L.~Chen, Y.~Wang, Ship recognition method combined with image
  segmentation and deep learning feature extraction in video surveillance,
  Multimedia Tools and Applications 79 (2020) 9177--9192.

\bibitem{ronneberger2015u}
O.~Ronneberger, P.~Fischer, T.~Brox, U-net: Convolutional networks for
  biomedical image segmentation, in: Medical Image Computing and
  Computer-Assisted Intervention--MICCAI 2015: 18th International Conference,
  Munich, Germany, October 5-9, 2015, Proceedings, Part III 18, Springer, 2015,
  pp. 234--241.

\bibitem{he2017mask}
K.~He, G.~Gkioxari, P.~Doll{\'a}r, R.~Girshick, Mask r-cnn, in: Proceedings of
  the IEEE international conference on computer vision, 2017, pp. 2961--2969.

\bibitem{ren2015faster}
S.~Ren, K.~He, R.~Girshick, J.~Sun, Faster r-cnn: Towards real-time object
  detection with region proposal networks, Advances in neural information
  processing systems 28 (2015).

\bibitem{chen2017deeplab}
L.-C. Chen, G.~Papandreou, I.~Kokkinos, K.~Murphy, A.~L. Yuille, Deeplab:
  Semantic image segmentation with deep convolutional nets, atrous convolution,
  and fully connected crfs, IEEE transactions on pattern analysis and machine
  intelligence 40~(4) (2017) 834--848.

\bibitem{zhao2017pyramid}
H.~Zhao, J.~Shi, X.~Qi, X.~Wang, J.~Jia, Pyramid scene parsing network, in:
  Proceedings of the IEEE conference on computer vision and pattern
  recognition, 2017, pp. 2881--2890.

\bibitem{zhang2021k}
W.~Zhang, J.~Pang, K.~Chen, C.~C. Loy, K-net: Towards unified image
  segmentation, Advances in Neural Information Processing Systems 34 (2021)
  10326--10338.

\bibitem{cheng2021per}
B.~Cheng, A.~Schwing, A.~Kirillov, Per-pixel classification is not all you need
  for semantic segmentation, Advances in Neural Information Processing Systems
  34 (2021) 17864--17875.

\bibitem{cheng2022masked}
B.~Cheng, I.~Misra, A.~G. Schwing, A.~Kirillov, R.~Girdhar, Masked-attention
  mask transformer for universal image segmentation, in: Proceedings of the
  IEEE/CVF Conference on Computer Vision and Pattern Recognition, 2022, pp.
  1290--1299.

\bibitem{tang2023can}
L.~Tang, H.~Xiao, B.~Li, Can sam segment anything? when sam meets camouflaged
  object detection, arXiv preprint arXiv:2304.04709 (2023).

\bibitem{ji2023segment}
W.~Ji, J.~Li, Q.~Bi, W.~Li, L.~Cheng, Segment anything is not always perfect:
  An investigation of sam on different real-world applications, arXiv preprint
  arXiv:2304.05750 (2023).

\bibitem{wei2022chain}
J.~Wei, X.~Wang, D.~Schuurmans, M.~Bosma, E.~Chi, Q.~Le, D.~Zhou, Chain of
  thought prompting elicits reasoning in large language models, arXiv preprint
  arXiv:2201.11903 (2022).

\bibitem{wang2023large}
Y.~Wang, Y.~Zhao, L.~Petzold, Are large language models ready for healthcare? a
  comparative study on clinical language understanding, arXiv preprint
  arXiv:2304.05368 (2023).

\bibitem{dosovitskiy2020image}
A.~Dosovitskiy, L.~Beyer, A.~Kolesnikov, D.~Weissenborn, X.~Zhai,
  T.~Unterthiner, M.~Dehghani, M.~Minderer, G.~Heigold, S.~Gelly, et~al., An
  image is worth 16x16 words: Transformers for image recognition at scale,
  arXiv preprint arXiv:2010.11929 (2020).

\bibitem{demir2018deepglobe}
I.~Demir, K.~Koperski, D.~Lindenbaum, G.~Pang, J.~Huang, S.~Basu, F.~Hughes,
  D.~Tuia, R.~Raskar, Deepglobe 2018: A challenge to parse the earth through
  satellite images, in: Proceedings of the IEEE Conference on Computer Vision
  and Pattern Recognition Workshops, 2018, pp. 172--181.

\bibitem{escobar2021satellite}
F.~Escobar, Satellite images of water bodies,
  \url{https://www.kaggle.com/datasets/lakshaymiddha/crack-segmentation-dataset},
  accessed: 2023-04-27 (2021).

\bibitem{al2020dataset}
W.~Al-Dhabyani, M.~Gomaa, H.~Khaled, A.~Fahmy, Dataset of breast ultrasound
  images, Data in brief 28 (2020) 104863.

\bibitem{rahman2021exploring}
T.~Rahman, A.~Khandakar, Y.~Qiblawey, A.~Tahir, S.~Kiranyaz, S.~B.~A. Kashem,
  M.~T. Islam, S.~Al~Maadeed, S.~M. Zughaier, M.~S. Khan, et~al., Exploring the
  effect of image enhancement techniques on covid-19 detection using chest
  x-ray images, Computers in biology and medicine 132 (2021) 104319.

\bibitem{ulucan2020large}
O.~Ulucan, D.~Karakaya, M.~Turkan, A large-scale dataset for fish segmentation
  and classification, in: 2020 Innovations in Intelligent Systems and
  Applications Conference (ASYU), IEEE, 2020, pp. 1--5.

\bibitem{diversis2021fire}
Fire segmentation image dataset,
  \url{https://www.kaggle.com/datasets/diversisai/fire-segmentation-image-dataset},
  accessed: 2023-04-27 (2021).

\bibitem{middha2021crack}
L.~Middha, Crack segmentation dataset,
  \url{https://www.kaggle.com/datasets/lakshaymiddha/crack-segmentation-dataset},
  accessed: 2023-04-27 (2021).

\bibitem{jafarian2021learning}
Y.~Jafarian, H.~S. Park, Learning high fidelity depths of dressed humans by
  watching social media dance videos, in: Proceedings of the IEEE/CVF
  Conference on Computer Vision and Pattern Recognition, 2021, pp.
  12753--12762.

\end{thebibliography}

%% else use the following coding to input the bibitems directly in the
%% TeX file.

\end{document}